# An Improved Intelligent Agent for Mining Real-Time Databases Using Modified Cortical Learning Algorithms

**N.E. Osegi**
Department of Information and Communication Technology
National Open University of Nigeria
Lagos State, Nigeria
E-mail: nd@osegi.com

## Abstract

Cortical Learning Algorithms based on the Hierarchical Temporal Memory (HTM) have been developed by Numenta Incorporation from which variations and modifications are currently being investigated upon. HTM offers better promises as a future computational model of the "neocortex" the seat of intelligence in the brain. Currently, intelligent agents are embedded in almost every modern day electronic system found in homes, offices and industries worldwide. In this paper, we present a first step in realising useful HTM like applications specifically for mining a synthetic and real-time dataset based on a novel intelligent agent framework, and demonstrate how a modified version of this very important computational technique will lead to improved recognition.

## I. Introduction

Cortical Learning Algorithms (CLA) are typically a suite of algorithms developed to implement some functionality of the mammalian neocortex in computer software. According to Numenta (Numenta, 2014), HTM is the computational theory on which CLA framework is built. Thus, a detailed understanding of HTM theory is important for any implementation of CLA. At the basic level, CLA is reduced to a cortical learning microcircuit or a sequence of intelligent cortical learning microcircuits. In this paper we developed a reduced version of CLA coined Reduced Cortical Learning Transphomers (rCLT) suitable for real-time embedded learning database systems. We put forward a new model of Cortical Learning based on the First-Last Rule (FLR) and the Frequent-Occurring Rules (FOR).

The paper is organized as follows:

In section II we briefly describe the HTM theory necessary for understanding the CLA-like algorithms. In section III we present the underlying concept of rCLT including the definition of a concept map based on two rule-concept sets. In section IV we experiment on a synthetic and real-time dataset as a proof-of-concept and present our results and discussions. We give our conclusions in section V.

## II. HTM Theory for Cortical Learning Microcircuits

Hierarchical Temporal Memory (HTM) is a computational theory of mammalian cortex that suggests the sparse distributed hierarchical learning of the brain over time. The aspect of time is very important since this guarantees that we are dealing with a living and dynamic being that is guaranteed to intelligently learn sparse patterns of the input world over time. HTM uses Approximate Computing techniques which tends to encourage nearest neighbour Associative Learning while keeping the learning objects (or elements) at a bare minimum. This form of expectation maximization has deep Bayesian probabilistic roots which is beyond the scope of this paper. The HTM theory tactically proposes four principles and four functions. It also suggests what each layer or region does in a cortical circuit. We shall briefly examine these principles, functions and proposals in this section.

### A. The HTM Principles

The principles (learning principles) include the following:

- The Use of a Hierarchy

- The Use of Regions

- The Use of a Sparse-Distributed Representation

- The Use of specific timing constraints

The hierarchy constitute an arrangement of HTM Regions which are memory elements organized in a columnar structure. The regions contain HTM cells which are actually random generative neurons with a spiking profile. The magnitude of the connectedness of these HTM neurons or cells will determine the direction of a winning or successful column. These regions represent the main units of memory and prediction in HTM.

The use of a sparse distributed structure ensures that at any point in time, the input to a HTM cortical circuit is a sparse representation of a hypothetical or real world sensor input.

### B. The HTM Functions

The HTM functions include the following:

- Learning

- Inference

- Prediction

- Behaviour

In the first stage (Spatial Pooling), learning is generically achieved through feed-forward pattern sequencing via the sparse distributed data structure and standard Hebbian (or Hebbian-Hopfield) updates using the notion of "Permanence" and "Boosting". Permanence defines the level and connectedness of a given sequence of HTM cells typically facilitated using synaptic points on the proximal or distal dendrites. Boosting is used to support weaker cells in the learning process. This HTM cells will in turn determine which set of overlapping columns will be used by the cortical circuit for online learning. This process is referred to as inhibition and the successful column(s) called "winner columns". Typically, the inputs to the spatial pooler are recurrent – continual sensor signals.

In the second stage we perform a temporal pooling operation on the output stage of the spatial pooler. This is achieved using a predictive sequence operation on HTM segments (group of cells) at time step, t.

The predicted sequences can then be used for making inference by matching novel or previously learnt inputs to the recognition or memory prediction units. It is important to note that at each point of the learning and prediction process, the data is sparse resulting in significant savings in memory in addition to higher representational efficiencies.

### C. Proposals for the HTM Cortical Layers or Regions

The HTM Layers include four key proposals. All layers are assumed to be feed-forward layers learning sequences of sequences of data.

- **V4-** These are Layer 4 cells. When these layers are present in a cortical circuit, they use the HTM-CLA to learn first-order predictions (FOP's) or temporal transitions. They make representations that are invariant to spatial transformation.

- **V3-** These are layer 3 cells. They are closest to HTM-CLA described in (Numenta, 2014). They use the HTM-CLA to learn variable-order prediction (VOP) or temporal transitions. They form stable representations that are propagated up the cortical hierarchy.

- **V5-** These are layer 5 cells. They learn VOP's with timing. They also possess motor/gating ability (see reticular formation).

- **V2-** These are layer 2 and layer 6 cells. Though no specific proposals are made, they are assumed to learn some form of sequence memory

### III. Reduced Cortical Learning Transphomer (rCLT)

The rCLT is a scaled down version of the CLA framework specifically targeted at embedded system applications. It follows a systematic functional/object-oriented procedure that facilitates easy debugging and refinements in software architecture. The ideas of rCLT is guided by the SDR theory in (Ahmad and Hawkins, 2015) and by the Generative Models in (Osegi and Enyindah, 2015)

## A. The rCLT Algorithm

The algorithm presumes that given a group or matrix of real world analogue sensor inputs, the sparse distributed representation in a cortical circuit is a random function of the most frequently occurring analogue inputs or the first-or-last occurring analogue inputs in the input sensor sequence.

Stated in Mathematical terms,

$$x_{sparse} > (x_f \; Or \; (x_{first} \; Or \; x_{last})$$
$$\text{where,}$$
$$x_f, x_{first}, and \; x_{last} \in \Re \quad (1)$$
$$x_{first}, and \; x_{last} \in U(0, ro)$$

A typical rule procedure is outlined in Appendix I and Appendix II for the FLR and FOS rules respectively, but is not intended to be limiting in this context. Source code implementations can be obtained from the Matlab File Exchange Website (www.matlabcentral.com).

## B. Spatial Pooling in rCLT

The spatial pooler algorithm forms a sparse distributed representation of the input sensing world. The spatial pooler implements some important cortical functions which are vital to the smooth and reliable operation of a cortical circuit. Spatial Pooling is implemented as follows:

Step. 1. Initialise all parameters and constants:

$I_{m(ro,co)}$

$K_o : resize \; constra \text{int}$

$k1 : frequent \; and \; first-last \; rule \; parameters \, (x_f \; Or \; (x_{first} \; Or \; x_{last}))$

$k2 : Sparsity \; criterion - typically \; 0.1 \, to \, 0.2$

$K_p : default \; permanence \; threshold$

$K_s : default \; matching \; score \; at \; po\text{int} \; s \; for \; obtaining \; winning \; column(s)$

$K_{score}$ : *default matching score for obtaining winning column(s)*

$ro$ : *number of rows*

$co$ : *number of columns*

Step. 2.  Perform a first-order sparsity by resizing the input. This is typical when processing large images or audio signals -where the data is large we use only a subset of the data at a particular instance of time. This is typical of most brain sensory fields.

Step. 3.  Max-out the inputs and build the partition function:

Max-out:

$$I_{max} = \max(\max(Io)) \qquad (2)$$

Partition:

$$I_z = (\frac{I_o \cdot}{I_{max}}) \forall_{I_o} \qquad (3)$$

Step. 4.  Apply equation (1) to the partition function in step 3 to form the 2-nd order sparse representation (Binary SDRs).

Step. 5.  Reshape the SDRs formed in step 4 to generate a single-dimensional columnar sensor input for online cortical training

$$\begin{aligned} i_{zr} &= reshape(I_z, i_{product}, 1) \\ i_{product} &= ro * co \quad \{ro, co \in Z \end{aligned} \qquad (4)$$

Step. 6.  Generate random columnar cellular SDRs constrained by Kp:

$$\begin{aligned} &for\, i = 1 : sparse\_cols \\ &\quad i_g = rand(i_{product}, 1) > K_p \end{aligned} \qquad (5)$$

- Step. 7. Form the Union Set : logical sum of output of step 6 including the influence of noise

- Step. 8. Compute Overlap: logical sum of product of the Union set and the Input SDR

- Step. 9. Extract winning columns and update permanences

### C. Temporal Pooling in rCLT

Temporal pooling follows a simple set of operations similar to the spatial pooler but with a timing constraint i.e. the learning and predictive sequences are time based. The outputs from the Spatial Pooler form the inputs to the temporal pooler in a feed-forward fashion.

## IV. Experimental Details and Results

The experiments were performed on a standard PC with 2.0GHz processor and enough RAM/Hard-disk Memory space.

The experiments were conducted in two parts:

**Approach 1:**

This approach uses a synthetic small sample dataset to study the performance of the rCLT. The percentage efficiencies using Approach 1 are given in Table 1. Figures 1 to 4, show the graphical matched SDR response of the rCLT cortical circuit against the input SDR set. The influence of noise sparsity have not been studied.

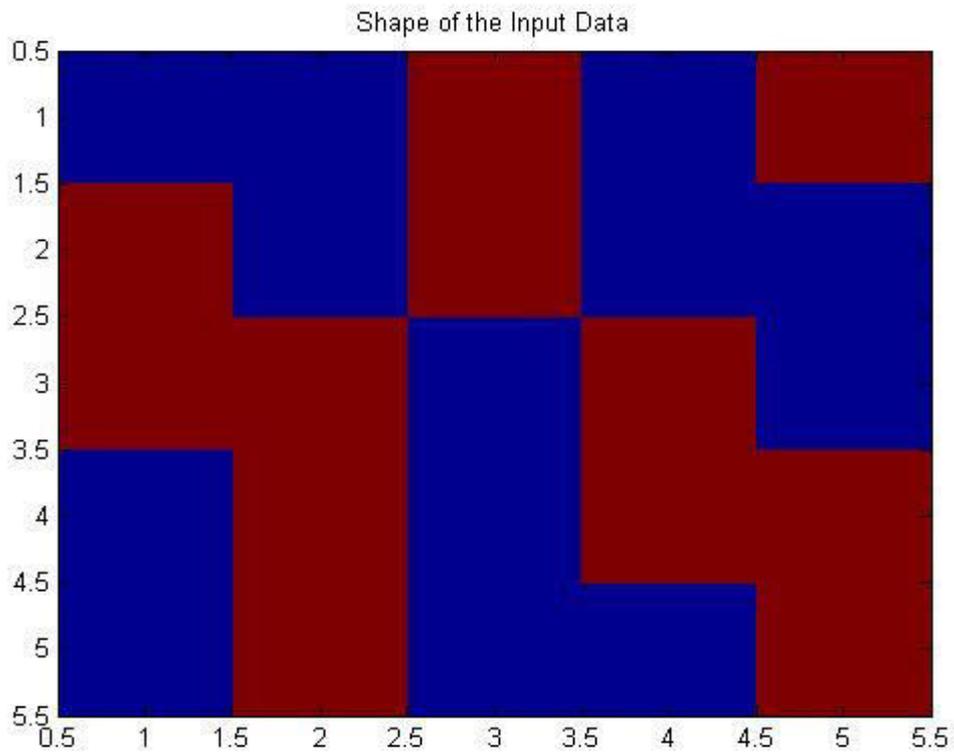

Fig1. Input SDR for Synthetic Dataset at time step, t = 4

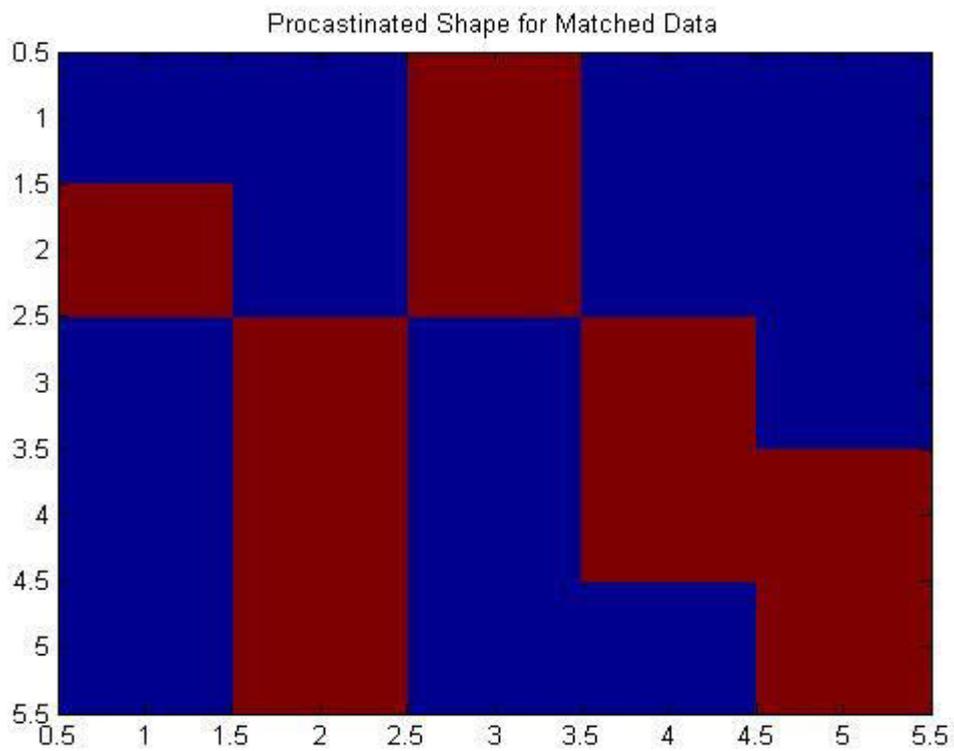

Fig2. Matched SDR for Synthetic Dataset at time step, t = 4

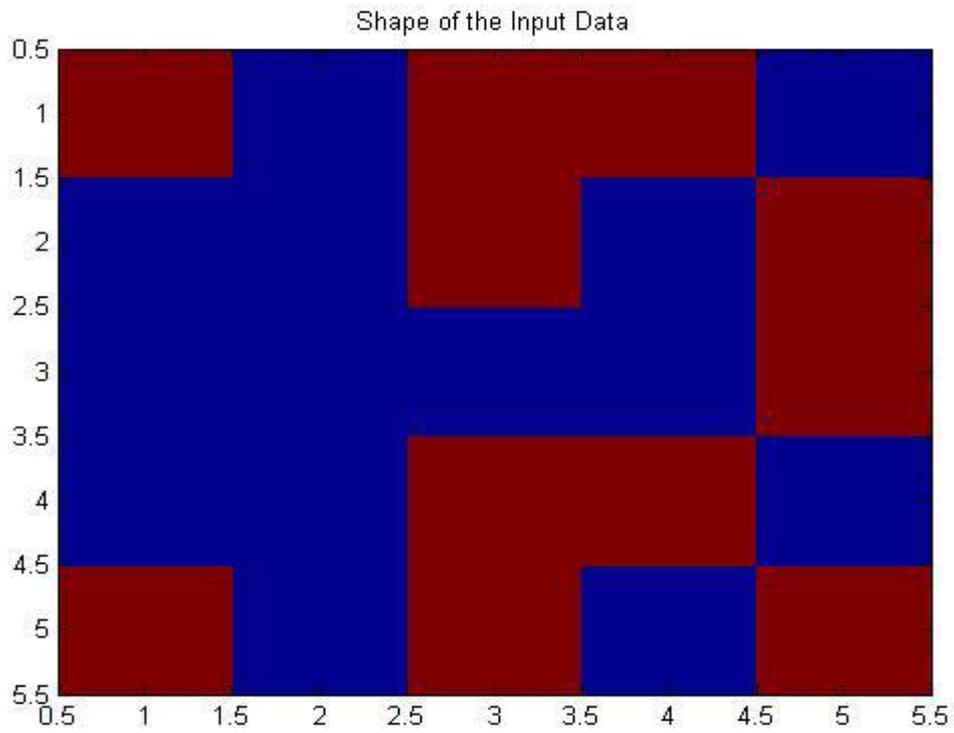

Fig3. Input SDR for Synthetic Dataset at time step, t = 5

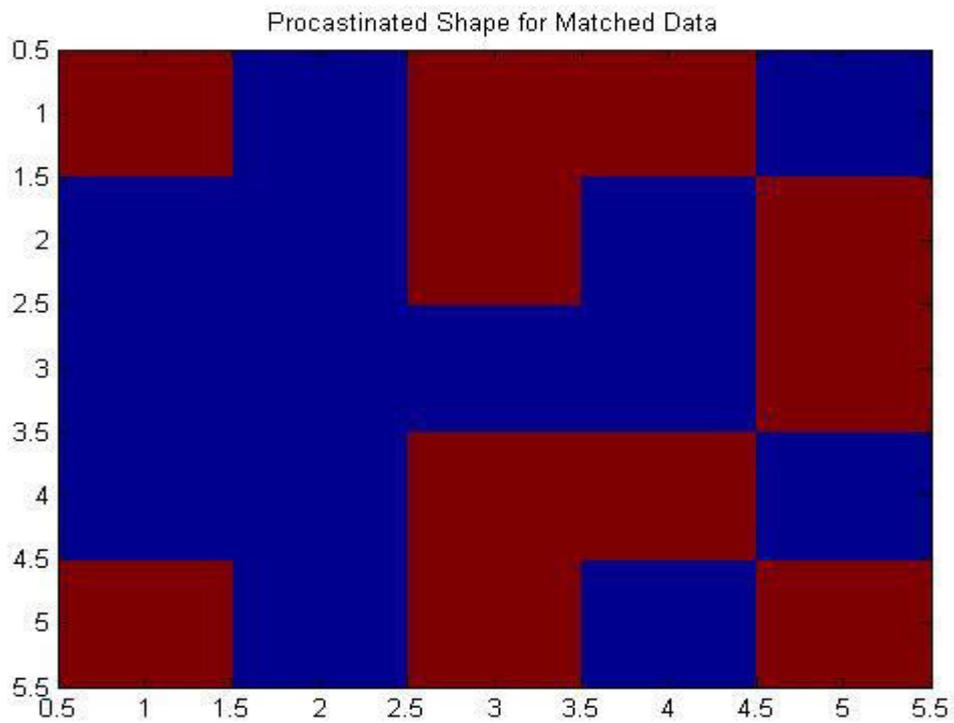

Fig4. Matched SDR for Synthetic Dataset at time step, t = 5

Table 1: Percentage Accuracies for the Synthetic dataset at c = 1

| Time step/Observation | % Accuracy |
|---|---|
| 1 | 100 |
| 2 | 100 |
| 3 | 100 |
| 4 | 92 |
| 5 | 100 |

**Approach 2:**

This approach uses a real-time streaming audio sample dataset to study the performance of the rCLT. The percentage efficiencies using Approach 1 are given in Table 2. Figures 5 to 8, show the graphical matched SDR response of the rCLT cortical circuit against the input SDR set. The influence of noise sparsity have not been studied.

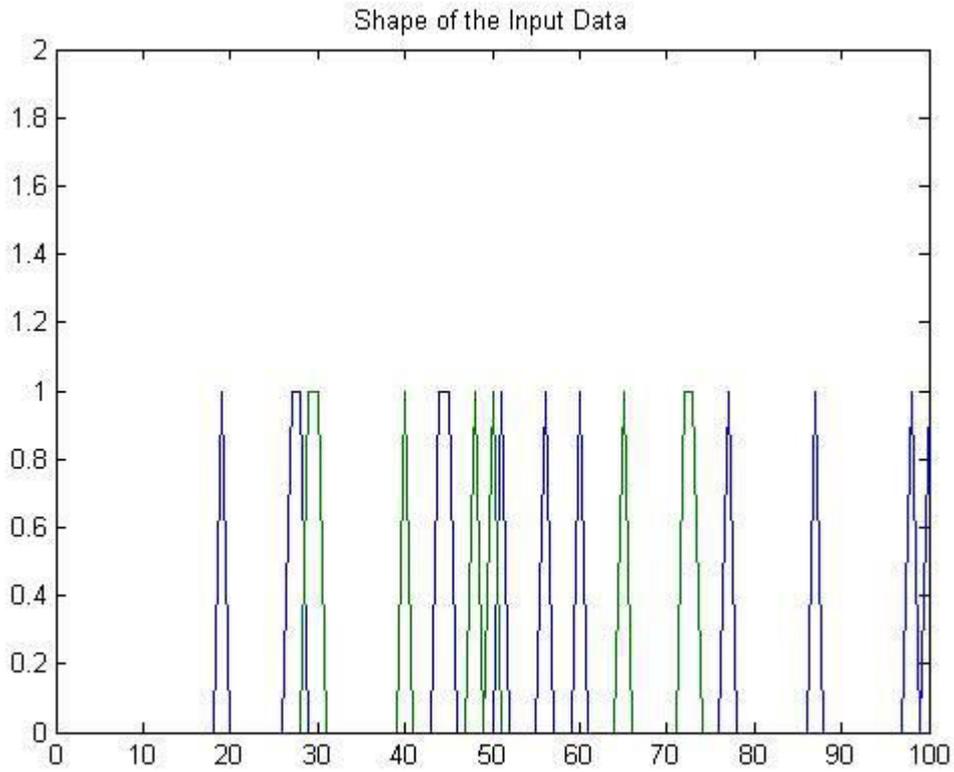

Fig5. Input SDR for Streaming Audio Dataset at time step, t = 1

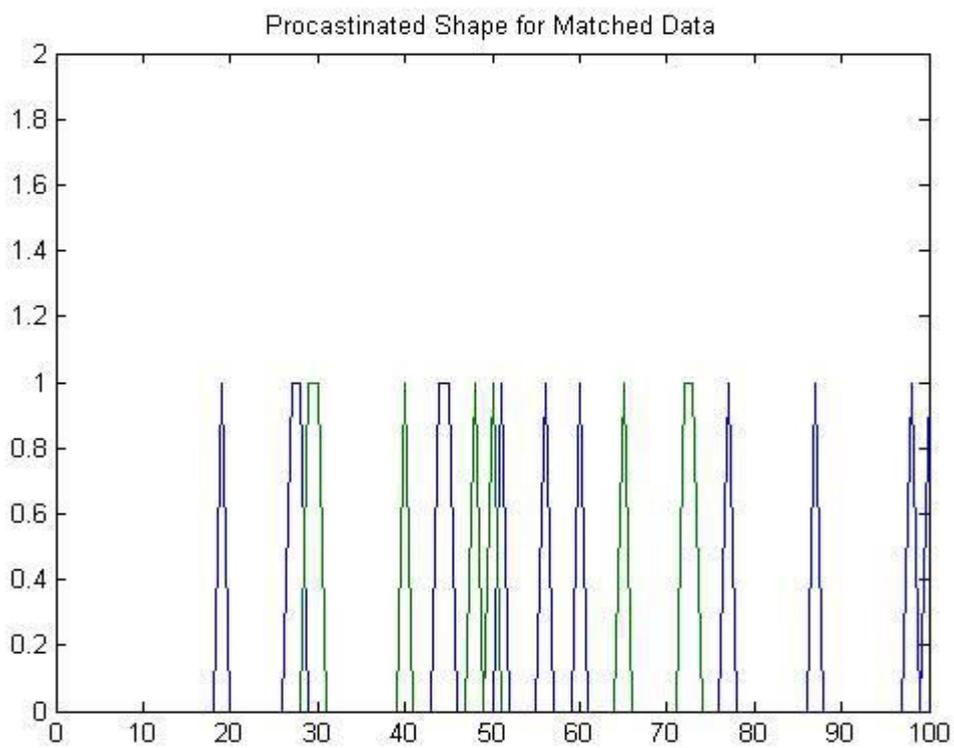

Fig6. Matched SDR for Streaming Audio Dataset at time step, t = 1

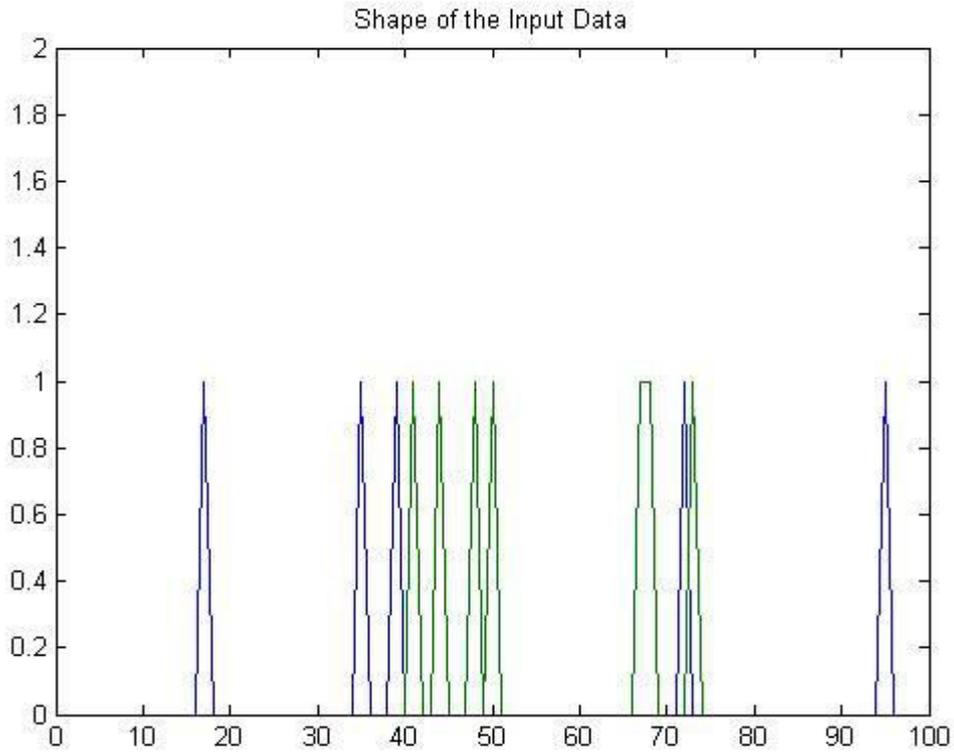

Fig7. Input SDR for Streaming Audio Dataset at time step, t = 5

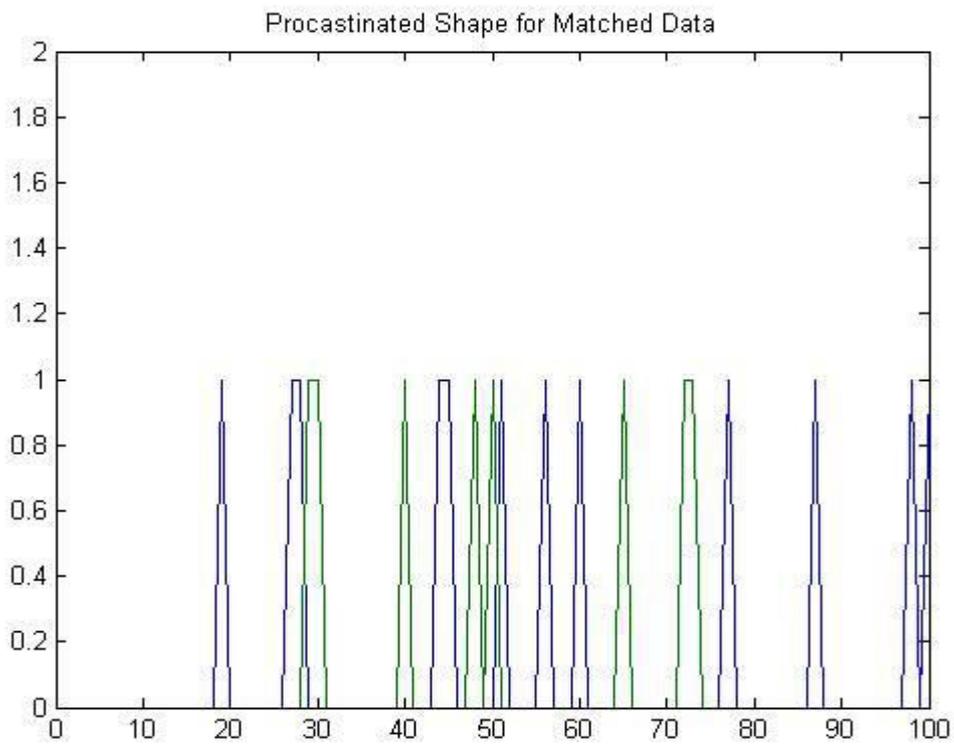

Fig8. Matched SDR for Streaming Audio Dataset at time step, t = 5

Table 2: Percentage Accuracies for the Streaming Audio dataset at c = 1

| Time step/Observation | % Accuracy |
|---|---|
| 1 | 100 |
| 2 | 99.5 |
| 3 | 99 |
| 4 | 99.5 |
| 5 | 99.5 |

From the results, it is easy to see that efficiencies greater than 92% is achievable using rCLT irrespective of the variation in input for only a single learning column. This can be greatly improved by adding more columns or when the inputs are repeatable.

## V. Conclusion

Cortical Learning Algorithms play a vital role in understanding how the mammalian neocortex operate but is still limiting in terms of functional strength and computational power when compared to real human cortex. In this paper, we have developed a novel intelligent agent framework based on a modified Cortical Learning Algorithm for embedded real-time database systems. The developed framework in this paper will provide a good starting point for developing structured object-oriented implementations of cortical learning microcircuits in real-time embedded applications.

# References

1. Hawkins, J., Ahmad, S., & Dubinsky, D. (2010). Hierarchical temporal memory including HTM cortical learning algorithms. *Techical report, Numenta, Inc, Palto Alto http://www. numenta. com/htmoverview/education/HTM_CorticalLearningAlgorithms. Pdf*

**2.** Ahmad, S., & Hawkins, J. (2015). Properties of Sparse Distributed Representations and their Application to Hierarchical Temporal Memory. *arXiv preprint arXiv:1503.07469*.

3. Osegi, N. E., & Enyindah, P. (2015). Learning Representations from Deep Networks Using Mode Synthesizers. *arXiv preprint arXiv:1506.07545*.

4. rCLTs: Reduced Cortical Learning Transphomers, *http://www.mathworks.com/matlabcentral/fileexchange/54713-reduced-cortical-learning-transphomers*

# Appendix I

# FIRST-LAST (F-L) RULE ALGORITHM

**Step. 1.**   Get Input

**Step. 2.**   Form SDR

**Step. 3.**   Generate Cells

**Step. 4.**   For Each Column Insert into Segments

**Step. 5.**   Get First Segment

**Step. 6.**   Get Last Segment

**Step. 7.**   Combine First and Last Segments to form F-L Synaptic Potentials

**Step. 8.**   Use F-L Synaptic Potential for Input Matching

**Step. 9.**   Store matched in memory to form memory-store

**Step. 10.**   Get New Inputs

**Step. 11.**   Use memory-store to form new predictions – against New Inputs

**Goal:** *To prove the Intermediate Manifold Hypothesis (IMF) – stated here as the possibility that there exist a high concentration of distributive information (i.e. the data generating distribution) at first and last point of a data sequence (s). The first and last points are assumed to be of low dimensionality.*

## Appendix II

## Frequently Occurring Segment (FOS) ALGORITHM

**Step. 12.** Get Input

**Step. 13.** Form SDR

**Step. 14.** Generate Cells

**Step. 15.** For each column Insert into Segments

**Step. 16.** Get Column with the most frequently occurring segments (FOS) – i.e. columns that give the highest number of similarly occurring segments

**Step. 17.** Use the most FOS to form Synaptic Potentials

**Step. 18.** Use FOS Synaptic Potential for Input Matching

**Step. 19.** Store matched in memory to form memory-store

**Step. 20.** Get New Inputs

**Step. 21.** Use memory-store to form new predictions – against New Inputs

**Goal:** *To validate the Mode Synthesizer Algorithm (MSA)*